\documentclass[letterpaper]{article} 
\usepackage{aaai25}  

\usepackage{times}  
\usepackage{helvet}  
\usepackage{courier}  
\usepackage[hyphens]{url}  
\usepackage{graphicx} 
\usepackage{natbib}  
\usepackage{caption} 
\usepackage{algorithm}
\usepackage{algorithmic}

\usepackage{booktabs} 
\usepackage{multirow}
\usepackage{array}
\usepackage{amsmath}
\usepackage{bbold}
\usepackage{svg}

\usepackage{newfloat}
\usepackage{listings}
\DeclareCaptionStyle{ruled}{labelfont=normalfont,labelsep=colon,strut=off} 
\floatstyle{ruled}
\newfloat{listing}{tb}{lst}{}
\floatname{listing}{Listing}
\usepackage{xcolor}  

\definecolor{soham}{rgb}{0.8, 0.2, 0.2}

\definecolor{blue}{rgb}{0.0, 0.5, 1.0}  

\definecolor{q}{rgb}{0.09, 0.61, 0.1} 

\definecolor{m}{rgb}{0.6, 0.33, 0.73}

\newcommand{\bigO}{\mathcal{O}}
 
\newcommand{\hmme}{\text{HMM-\textit{e}\ }}

\usepackage{subfiles}

\usepackage{xr}

\title{Ensemble Methods for Sequence Classification with Hidden Markov Models}
\author{
    Maxime Kawawa-Beaudan\equalcontrib,
    Srijan Sood\equalcontrib,
    \\
    Soham Palande,
    Ganapathy Mani,
    Tucker Balch,
    Manuela Veloso
}
\affiliations{
    J.P. Morgan AI Research\\
    383 Madison Avenue\\
    New York, NY USA\\
    \{maxime.kawawa-beaudan, srijan.sood, soham.palande, ganapathy.mani, tucker.balch, manuela.veloso\}@jpmchase.com
}

\begin{document}

\maketitle

\begin{abstract}

We present a lightweight approach to sequence classification using Ensemble Methods for Hidden Markov Models (HMMs). HMMs offer significant advantages in scenarios with imbalanced or smaller datasets due to their simplicity, interpretability, and efficiency. These models are particularly effective in domains such as finance and biology, where traditional methods struggle with high feature dimensionality and varied sequence lengths. Our ensemble-based scoring method enables the comparison of sequences of any length and improves performance on imbalanced datasets.

This study focuses on the binary classification problem, particularly in scenarios with data imbalance, where the negative class is the majority (e.g., normal data) and the positive class is the minority (e.g., anomalous data), often with extreme distribution skews. We propose a novel training approach for HMM Ensembles that generalizes to multi-class problems and supports classification and anomaly detection. Our method fits class-specific groups of diverse models using random data subsets, and compares likelihoods across classes to produce composite scores, achieving high average precisions and AUCs.

In addition, we compare our approach with neural network-based methods such as Convolutional Neural Networks (CNNs) and Long Short-Term Memory networks (LSTMs), highlighting the efficiency and robustness of HMMs in data-scarce environments. Motivated by real-world use cases, our method demonstrates robust performance across various benchmarks, offering a flexible framework for diverse applications.




\end{abstract}

\section{Introduction} \label{sec:introduction}

Sequence modeling is a fundamental task in machine learning, with wide-ranging applications across multiple domains. Sequences span many data types, from categorical tokens (e.g., natural language, biological markers), continuous values (e.g., temperature, sensor readings), and mixed multivariate data (e.g., trading activity, user behavior, system logs). Sequence modeling problems are relevant in healthcare (diagnosis from patient and test history, genome and protein classification \cite{healthcare_sequence_classification, Gresova2022}), finance (anomaly detection in sequences of financial activity like credit card fraud, money laundering, and anomalous trades \cite{hmm_card_fraud, Byrd_2022}), technology (monitoring system logs and sensor data \cite{intrusion_detection_review}), and operations (identifying system and pipeline failures \cite{duarte2024event}).


Despite considerable progress in sequence classification learning methods, challenges persist, particularly with high-dimensional data, varying sequence lengths, the demand for real-time processing, and, most importantly, data imbalances present in real-world datasets. Existing approaches broadly fall into two main categories: traditional methods and modern deep learning-based models \cite{categorization_sequence_classification_models}. Traditional methods offer interpretability but often fail to capture the intricate temporal dependencies in sequential data. Conversely, modern deep learning methods like RNNs, LSTMs, and Transformers excel at modeling complex sequences \cite{kansara2020traditionalvsdeeplearning} but face issues like high computational costs, overfitting, and reduced interpretability. Both sets of approaches are significantly limited by severe data imbalances often seen in real-world data streams.

Typically, one class (usually the positive class) is underrepresented, while the other (negative class) dominates the dataset. For example, fraudulent transactions are far less common than legitimate ones, normal operations vastly outnumber anomalous events in systems, and bot activity is a fraction of normal user activity. Class imbalance leads to biased models that perform well on the negative majority class, but poorly on the positive class (failing to detect the rare but crucial positive instances) \cite{he2009_imbalanceddatageneral}. 

Hidden Markov Models (HMMs) are an established technique for sequence modeling, known for their ability to capture temporal dependencies and handle variable sequence length. They are particularly well suited in domains that require interoperability, and can be integrated with modern deep learning techniques for enhanced performance \cite{hmms_deeplearning_integration}. However, traditional HMMs also struggle with class imbalance. 

We propose a novel approach that leverages ensembles of HMMs for the task of sequence classification, particularly in scenarios with extreme class imbalance (such as anomaly detection), while remaining extendable to multi-class classification. While our real-world deployment scales to millions of sequences, we demonstrate our approach on established genome classification benchmark tasks.

Our contributions are as follows:
\begin{enumerate}
\item We propose a novel framework leveraging HMM ensembles (HMM-\textit{e}) for sequence classification, and establish training and inference procedures.
\item We propose a novel method for constructing composite scores from base learners in an ensemble, allowing for the comparison of sequences of extremely different lengths. This method is model-agnostic, and can be used with ensembles of other model classes.
\item We demonstrate the robustness of \hmme to class imbalances. 
\item We demonstrate the utility of \hmme as feature extractors for downstream classifiers, and show that the combination of HMM ensembles with downsteam models (SVMs and neural networks) outperforms established deep-learning baselines.
\end{enumerate}
 \label{sec:introduction}

\section{Background \& Related Work} \label{sec:background}

\subsection{Sequence Classification}
Sequence classification, the categorization of sequential data into semantically meaningful classes, is an extensively researched problem in machine learning, essential for tasks across various domains \cite{xing2010brief}.

\subsubsection{Data Imbalance and Anomaly Detection}

Many real-world problems involve the detection of rare events like system intrusions \cite{intrusion_detection_review}, credit card fraud \cite{hmm_card_fraud}, or money laundering \cite{jpm_air_aml}. These anomaly detection problems are made difficult by class imbalance, or the scarcity of examples for the behavior of interest. 

One-class anomaly detection approaches avoid modeling the anomalous class, as its examples are so sparse, and focus instead on robustly modeling the nominal behavior class. They use these models to measure the distance between observed and nominal behaviors. 
In other applications, such as those motivating our use case, both the normal and anomalous classes are modelled. This more targeted approach allows us to detect only particular kinds of behavioral anomalies, not just any behavior deviating from nominal. 

Synthetic data approaches can also be used in conjunction with other methods to augment the data of the underrepresented class \cite{dahmen2019synsys, rotem2022transfer, potluru2023synthetic}. While synthetic data is not the focus of our study, our approach also inherently provides a mechanism to generate synthetic samples.

\subsection{Hidden Markov Models (HMMs)} \label{sec:background_hmms}

Hidden Markov Models (HMMs) are statistical models for sequential data, which have a long history of use in natural language processing, finance, and bioinformatics \cite{hmm_intro_1986, hmm_biology_recent_survey, hmm_genomics_survey, hmm_stock_selection}. While they are simpler than many deep-learning based models, they can still robustly model sequential data by making the assumption that an underlying observation sequence $\bigO = \{o_1, o_2, ..., o_T\}$ is a series of unobserved transitions between hidden states $\{a_1, a_2, ..., a_T\}$. HMMs make two fundamental assumptions:
\begin{equation}
\label{eqn:markov_states}
p(A_{t+1}=a_{t+1} \mid a_t,..., a_1) = p(A_{t+1}=a_{t+1} \mid a_t) 
\end{equation}
\begin{equation}
\label{eqn:emission_independence}
p(O_t=o_t \mid a_t, o_{t-1}, a_{t-1},...,o_1, a_1)=p(O_t=o_t \mid a_t)
\end{equation}

Eq.~\ref{eqn:markov_states} describes the Markov property of the hidden states: the next state is independent of the rest of the state history given the curent state. Eq. \ref{eqn:emission_independence} describes the conditional independence of the current observation from the rest of the state and observation history given the current state. 
 
An HMM with $n$ states and a vocabulary of $m$ possible emissions consists of three parameter sets: the initial state distribution $\pi$, the $n \times n$ transition matrix $A$, and the $n \times m$ emission distribution $B$. These parameters are collectively denoted by $\lambda = (A, B, \pi)$. There are three fundamental HMM problems:

\begin{enumerate}
    \item Estimating the likelihood $p(\bigO \mid \lambda)$ of an observation sequence $\bigO$ given an HMM with parameters $\lambda$
    \item Estimating the most likely state sequence $(a_1,...,a_T)$ given an observation sequence $(o_1,...,o_N)$.
    \item Estimating the parameters $\lambda$ given an observation sequence $\bigO$ or a set of observation sequences $\bigO_1^N$ to maximize $p(\bigO \mid \lambda)$ or $p(\bigO_1^N \mid \lambda)$
\end{enumerate}

There exist long-standing solutions for each of these problems: problem 1 is solved via the forward-backward algorithm, problem 2 via the Viterbi Algorithm, and problem 3 via the Baum-Welch algorithm \cite{hmm_intro_1986}.

While many state-of-the-art approaches for sequential data problems have adopted deep-learning based models like transformers \cite{attention_is_all_you_need}, recent works have shown the applicability of HMMs to time series analysis \cite{hmm_ensemble_time_series}, financial applications like stock selection and trading \cite{hmm_stock_selection, hmm_stock_trading}, and behavioral anomaly detection in robotics \cite{azzalini_hmm_ad}. Although HMMs have also been leveraged in ensemble settings \cite{rezek2005ensemble, hmm_ensemble_time_series}, we're not aware of any concrete frameworks that establish training and inference routines for sequence classification.

\section{Approach}
\subsubsection{Singleton HMMs}
We first investigate the use of single HMMs in (binary) sequence classification. In this setup, the training data is separated by class, and each class' training sequences are used to train individual HMMs, yielding one positive class HMM $\lambda^+$ and one negative class HMM $\lambda^-$. For a new unseen sequence, the predicted label is the class of the HMM that assigns the sequence the higher likelihood:
\begin{equation}
\label{eqn:max_likelihood_classifier}
c(\bigO) = \mathbb{1}\{p(\bigO \mid \lambda^+) > p(\bigO \mid \lambda^-)\}
\end{equation}

\subsubsection{Variable Sequence Length}\label{subsection_variable_sequence_length}
HMMs have been applied to sequence analysis with tremendous success \cite{hmm_biology_recent_survey}. While obtaining a sequence's likelihood under a trained HMM is straightforward, comparing the likelihoods of sequences of varying lengths poses a significant challenge (despite length-based normalization approaches). Sequence length heavily influences the likelihood due to sequential multiplication of probabilities at each timestep. Some deep-learning based approaches are often similarly limited by the need to size parameter matrices to fixed lengths a priori, resulting in a constrained maximum sequence length.

We leverage model-driven normalization by computing a set of likelihood scores for a given sequence across multiple models and using these scores to determine an overall rank or score, rather than directly comparing sequence likelihoods.
\begin{figure*}[tb]
\centering
\includegraphics[width=0.9\linewidth]{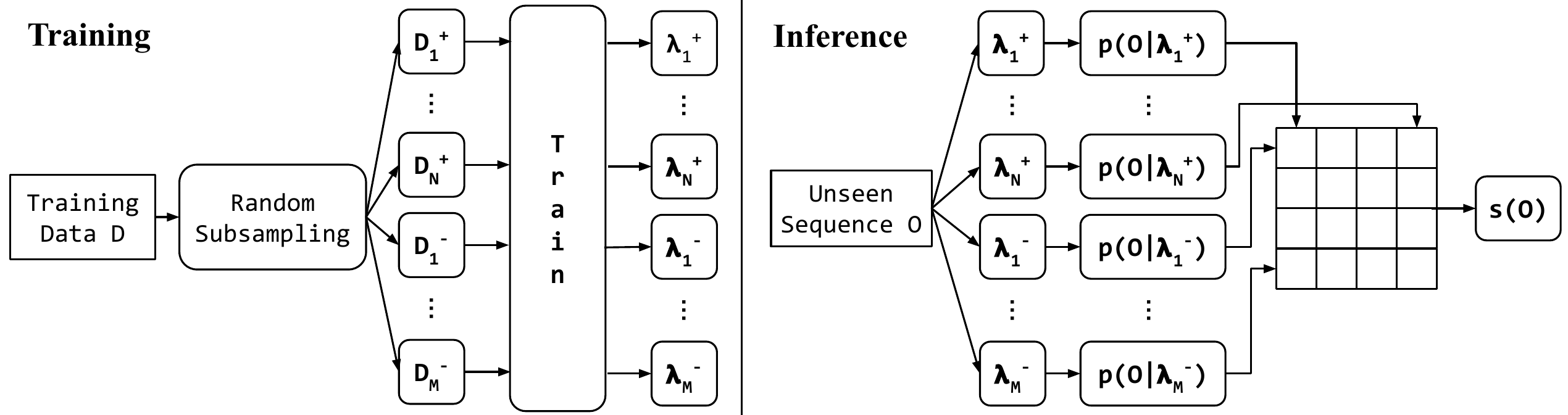} 
\caption{Flow diagram of our \hmme ensemble training and inference approach, as detailed in Section \ref{sec:hmm_ensembles}. While we adopt this approach using HMMs, the framework itself is model agnostic. The training data is broken into random subsets, and a diverse ensemble of learners is trained on these subsets. At inference time, pairwise matchups of likelihoods given by the models are compared, giving the composite score $s$.}
\label{fig:approach_diagram}
\end{figure*}


\subsection{HMM Ensembles}\label{sec:hmm_ensembles}
HMMs, while relatively simple and lightweight, can struggle to capture the complexity of behaviors represented in training data, especially when trained as a single model per class. Ensemble frameworks involve training multiple models over subsets of the training data \cite{dietterich2000ensemble}. This allows each learner to specialize in capturing distinct patterns or behaviors, while still modeling the entire data distribution collectively, leading to a more robust and generalize approach. Ensemble-based approaches also do well in scenarios with data imbalance, whereas monolithic models skew towards modeling the majority class (or underfitting in case of class-specific models) \cite{kuncheva2014combining}.

We propose \textbf{HMM-\textit{e}}, an ensemble framework for computing composite scores from the individual learner scores in our ensemble. This framework is model-agnostic, and while we find that HMMs work well for our use case, the base learners could just as well be deep-learning based models, SVMs, decision trees, or any other model class.

\subsubsection{Formalization and Algorithmic Framework}\label{sec:formalization}
First, we train $N$ models $\{\lambda_1^+,...,\lambda_N^+\}$ on the positive class and $M$ models $\{\lambda_1^-,...,\lambda_M^-\}$ on the negative class, taking care to ensure diversity among the models by training each on a randomly selected subset of samples from the training data. Each model sees $s\%$ of the training data in its relevant class. While we set $N = M$ for all settings, these parameters ($M, N, s$) can be established using typical hyperparameter optimization approaches. For any given sequence in the training data, the probability of not being selected for any model's random subset is $(1-s)^N$. The expected number of unsampled sequences is the same, so it is important to select $s$ and $N$ to keep this proportion of the training data small. 

For a previously unseen observation sequence $\bigO$, we compute the likelihood scores of  the sequence under all models: $\{p(\bigO \mid \lambda_1^+),...,p(\bigO \mid \lambda_N^+)\}$ and $\{p(\bigO \mid \lambda_1^-),...,p(\bigO \mid \lambda_M^-)\}$. We then compute the composite score:

\begin{equation}
\label{eqn:composite_score}
s(\bigO) = \sum_{i=1}^N \sum_{j=1}^M \mathbb{1} \{p(\bigO \mid \lambda_i^+) > p(\bigO \mid \lambda_j^-) \}  
\end{equation}
The composite score represents the number of unique pairwise matchups of positive class models and negative class models in which the positive class model assigns the sequence a higher likelihood than the negative class model.

\begin{figure}[htb]
\centering
\includegraphics[width=0.5\textwidth]{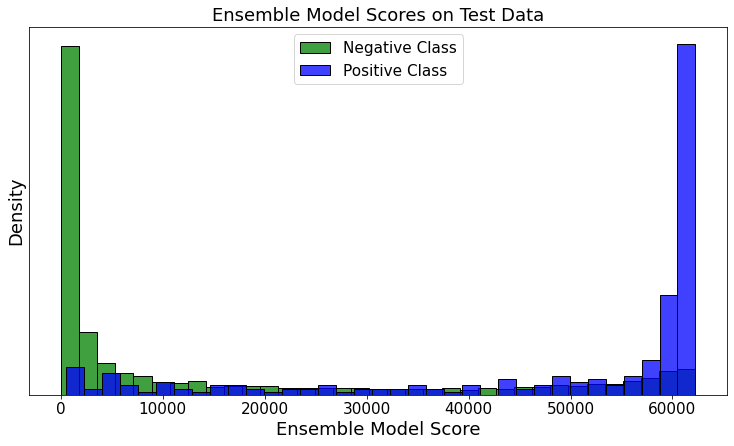}
\caption{The distribution of composite scores for test data in \textit{demo\_coding\_vs\_intergenomic\_seqs}, using a 250-model ensemble in an imbalanced data setting with class ratio 50:1. We observe good class separation even with data imbalance.} \label{fig:score_distribution}
\end{figure}

$s(\bigO)$ therefore takes on values in the range $[0, N \times M]$, where $s(\bigO) = 0$ indicates that negative class models outscored positive class models in all pairwise matchups, and $s(\bigO) = N\times M$ the opposite. The lower $s(\bigO)$ is, the less likely a sequence is to be in the positive class, and vice versa. A sample distribution of composite scores over a corpus of held-out sequences, colored by class, is shown in Fig. \ref{fig:score_distribution}. We see that the classes are well-separated, and indeed a decent classifier can be produced simply by thresholding the distribution at around $s(\bigO) = 50,000$.

Given the formulation in Eq. \ref{eqn:composite_score}, likelihoods for sequences of different lengths are never compared. Rather, given a corpus of sequences $\{\bigO_1,...,\bigO_S\}$ and the distribution of scores $\{s(\bigO_1),...,s(\bigO_S)\}$, sequences can be compared and classified on the basis of $s(\bigO)$. Eq. \ref{eqn:composite_score} can be thought of as an approach to normalizing for sequence lengths. $N$ and $M$ should be chosen such that the range $[0, N \times M]$ is sufficiently wide to differentiate the classes.

\begin{algorithm}[htb]
\label{alg:algorithm}
\caption{Ensemble Training and Inference}
\begin{algorithmic}[1]
\REQUIRE Training data $D$, number of positive models $N$, number of negative models $M$, subset factor $s$
\\
\hspace*{-2\algorithmicindent} \hrulefill\\
 \hspace*{-\algorithmicindent} \textbf{Training Phase} \\
 \hspace*{-2\algorithmicindent} \hrulefill
\STATE Randomly split $D$ into subsets $\{D_1^+, D_2^+, \ldots, D_N^+\}$ and $\{D_1^-, D_2^-, \ldots, D_M^-\}$ such that each subset contains $s\%$ of the relevant class data.
\FOR{$i = 1$ to $N$}
    \STATE $\lambda_i^+ \gets \text{train\_model}(D_i^+)$
\ENDFOR
\FOR{$j = 1$ to $M$}
    \STATE $\lambda_j^- \gets \text{train\_model}(D_j^-)$
\ENDFOR

 \hspace*{-2\algorithmicindent} \hrulefill \\
 \hspace*{-\algorithmicindent} \textbf{Inference Phase}
\\\hspace*{-2\algorithmicindent} \hrulefill
\setcounter{ALC@line}{0} 
\REQUIRE Corpus of sequences $C = \{\bigO_1,...,\bigO_S\}$
\STATE Initialize an empty list of composite scores $S$
\FOR{each sequence $\bigO_i \in C$}
    \STATE Compute likelihood scores $\{p(\bigO_i \mid \lambda_1^+), \ldots, p(\bigO_i \mid \lambda_N^+)\}$ and $\{p(\bigO_i \mid \lambda_1^-), \ldots, p(\bigO_i \mid \lambda_M^-)\}$
    \STATE Compute composite score $s(\bigO_i)$ and append to $S$
\ENDFOR
\STATE Determine threshold $s_{thresh}$ based on scores in $S$
\FOR{each sequence $\bigO_i \in C$}
    \STATE Class $c(\bigO_i) = \mathbb{1}\{ s(\bigO_i) \geq s_{thresh} \}$
\ENDFOR
\end{algorithmic}
\end{algorithm}

\subsubsection{Model Diversity in Ensembles}

\begin{figure*}[htb]
\centering
\includegraphics[width=0.78\linewidth]{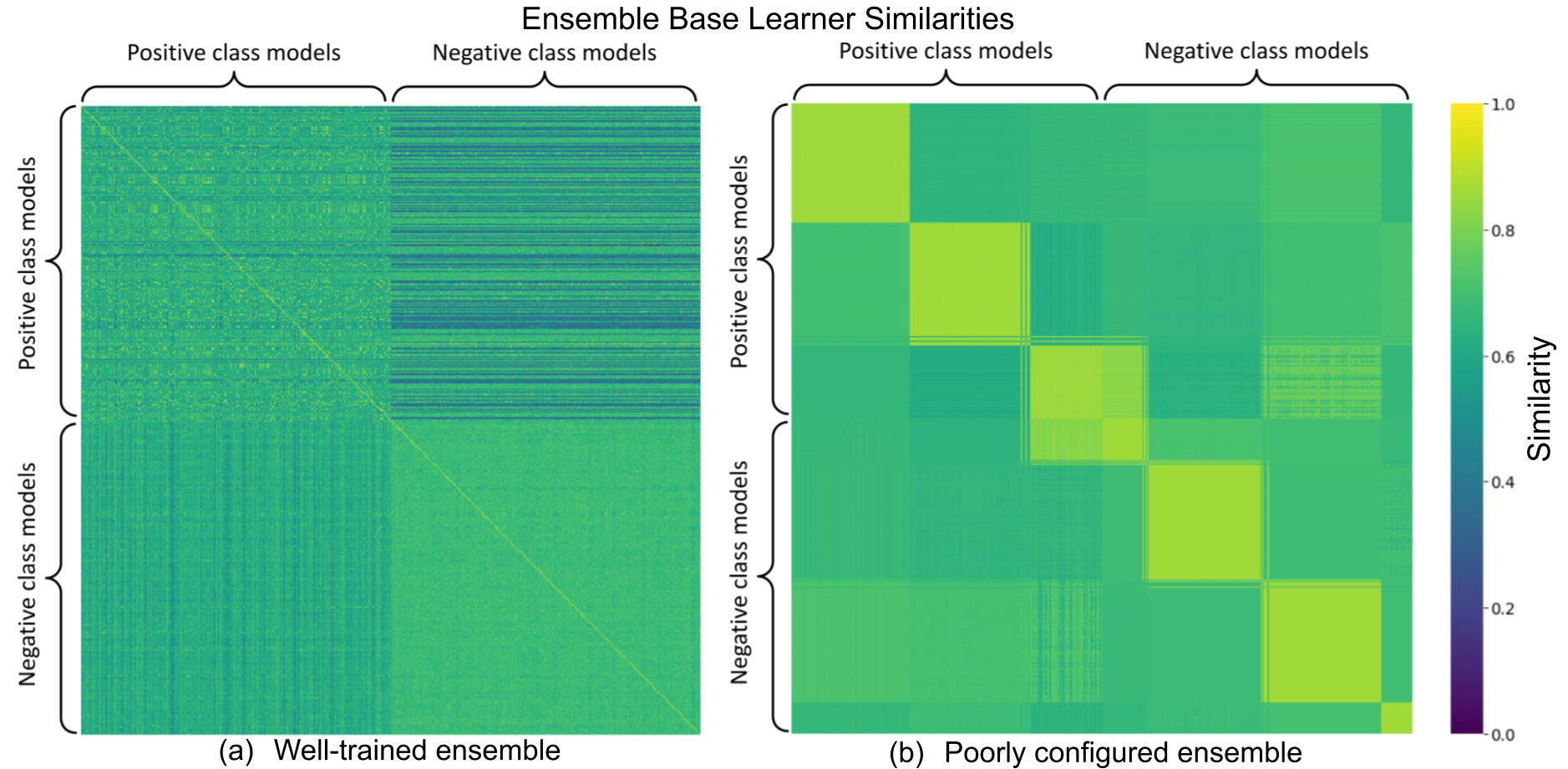}
\caption{Pairwise inter-HMM similarities for models in the 250-model ensemble on dataset \textit{demo\_human\_or\_worm}. In the properly configured setting (left) models are trained from unique initializations for 25 iterations. In the degenerate setting (right) subsets of models are trained from identical initializations for 5 iterations. We see that without sufficient diversity the ensemble may learn redundant models, effectively reducing the ensemble size.}
\label{fig:hmm_similarity_degenerate}
\end{figure*}

It is important to ensure the diversity of the trained models in any ensembling approach. Without sufficient diversity amongst training data subsets or model parameters, ensembles may learn redundant models. This reduces the efficiency of the ensemble due to redundancies.
We demonstrate this in Fig. \ref{fig:hmm_similarity_degenerate} by comparing sub-model similarities in two settings: (a) a well trained ensemble (right), and (b) a poorly initialized ensemble (left). For each unique pair of HMMs in the ensemble we plot the HMM similarity $1 - D(\lambda_i, \lambda_j)$, where $D(\lambda_i, \lambda_j)$ is the HMM distance proposed in \cite{azzalini_hmm_ad}. This distance is a sum of Hellinger distances between emission distributions of states matched using linear sum assignment, and weighted by the stationary distribution among states.

Fig. \ref{fig:hmm_similarity_degenerate}(a) demonstrates the scenario where \hmme is properly trained. As expected, the dark diagonal demonstrates that $D(\lambda_i, \lambda_i) = 0 \; \forall i$, while the noisy off-diagonals  demonstrate non-zero and uncorrelated distances for $D(\lambda_i, \lambda_j) \; \forall i,j, \; i \neq j$. We see four large blocks showing that intra-class HMM similarities are higher on average the inter-class similarities, as desired.

Fig. \ref{fig:hmm_similarity_degenerate}(b) demonstrates the effect of reduced ensemble sub-model diversity. This is done by artificially inducing the learning of redundant models on \textit{demo\_human\_or\_worm}. HMMs are sensitive to their initialization, and we seed subsets of the ensemble with the same initialization. We also train for only 5 iterations (vs 25) to increase the sensitivity to initialization. The blocks of similarly-colored HMM pairs show that many models are redundant. Each of these blocks can be thought of as one highly inefficient meta-model, effectively reducing the ensemble size to the number of blocks. 

Ensemble size $N$ and subset factor $s$ are important hyperparameters for ensuring diversity -- if $N$ is too small, our coverage of behaviors in the training data is poor. If $s$ is too large, we learn redundant models. This is discussed in Section~\ref{sec:ensemble_parameters}.
Individual learner parameters can also be modulated over the ensemble to encourage diversity. For HMMs, the number of states $n$ is the most impactful hyperparameter, and we observe some improvement ($\sim$3\% AUC-ROC) by varying the number of states $\in [3, 4, 5]$ across models rather than using $n=5$ for all models. In general, more diverse ensembles are more efficient as they can achieve equivalent performance to less diverse ensembles at smaller ensemble sizes.

\subsubsection{Generative Properties of HMMs}
Well-trained HMMs possess inherent generative capabilities, and can be used to generate synthetic data based on the properties of the original training data~\cite{ferrando2018generation, dahmen2019synsys}. This is particularly valuable in scenarios with class imbalance or data scarcity, and can be done with both singleton and ensemble settings. Although we do not explore this direction in this study, it is a promising avenue for future research.

Though our use case involves categorical, single-feature data, none of the HMM approaches discussed above are restricted to the categorical or single-feature cases. Both the singleton and \hmme approaches can be applied to continuous data using Gaussian HMMs or Gaussian Mixture Model HMMs, and multifeature data without any model changes necessary.

\subsection{Downstream Modeling using \hmme Scores}

Given a corpus of sequences $\{\bigO_1,...,\bigO_S\}$ and the distribution of scores $\{s(\bigO_1),...,s(\bigO_S)\}$, we can set a threshold $s_{thresh}$ and classify sequences directly as:

\begin{equation}
\label{eqn:ensemble_classification_rule}
c(\bigO_i) = \mathbb{1}\{ s(\bigO_i) \geq s_{thresh} \}
\end{equation}

We can also use the likelihoods of the base learners as features for consumption by downstream classifiers. For each sequence $\bigO_i$, we construct a feature vector comprised of the \hmme scores,
 \begin{equation}
 \label{eqn:feature_vector_definition}
        f_i = \begin{bmatrix}
           p(\bigO_i | \lambda_1^+),
           \hdots, 
           p(\bigO_i | \lambda_N^+),
          p(\bigO_i | \lambda_1^-),
           \hdots,
           p(\bigO_i | \lambda_M^-)
         \end{bmatrix}
  \end{equation}
Given that the magnitude of $f_i$ is highly sensitive to the sequence length, as discussed in Section~\ref{subsection_variable_sequence_length}, we normalize $f_i$ by $\lvert\lvert  f_i \rvert\rvert_2$ before training classifiers on $\{f_1,...,f_S\}$.

\begin{figure}[htb!]
\centering
\includegraphics[width=0.45\textwidth]{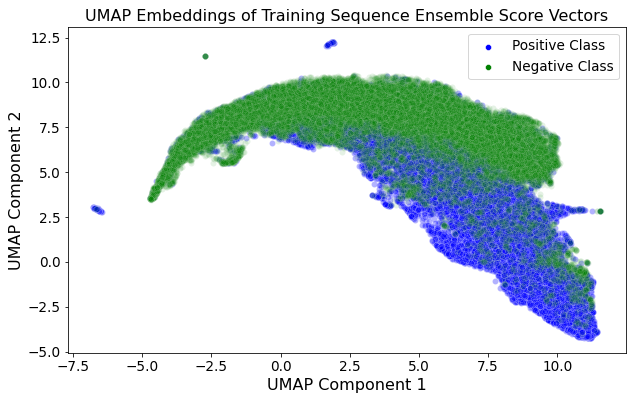}
\caption{UMAP embeddings of training data from dataset \textit{demo\_human\_or\_worm}, using the likelihood vectors from \hmme containing 250 sub-models for each class.}
\label{fig:umap}
\end{figure}

This technique utilizes HMMs as feature extractors, where each feature $p(\bigO_i | \lambda_j)$ represents the similarity between the sequence $\bigO_i$, and the random subset of training data underlying $\lambda_j$. Fig. \ref{fig:umap} shows the Uniform Manifold Approximation and Projection (UMAP) projections of these features $f_i$ for the dataset \textit{demo\_human\_or\_worm}, with the positive class in blue and the negative class in green. This unsupervised dimensionality reduction highlights the class structure, suggesting that the full high-dimensional feature vectors may be linearly separable. We train two types of classifiers in this framework: support vector machines (\textbf{\hmme + SVM}) and simple Neural Networks (\textbf{\hmme + NN}).

\section{Experiments}

\subsection{Data} \label{sec:data}
Our work is applicable to many real-world tasks in domains spanning healthcare, finance, operations, and technology. To demonstrate our approach's effectiveness, we leverage an open-source genomics benchmark \cite{Gresova2022} that consists of eight datasets for different classification problems. The benchmark represents our real-world problem setting well: each dataset contains sequences of tokens from a small vocabulary representing the four nucleotide bases (A, T, C, and G) and an unknown base token (N). Labels are provided at the sequence level. Sequences are varied in length and contain genetic information from a diverse set of model organisms including humans, mice, and roundworms. Labels correspond to the function and behavior of each genomic sequence -- whether the sequence is a promoter gene, enhancer gene, or an open chromatin region.  We provide results on four datasets included in the benchmark. 

All datasets are split by the benchmark authors into train and test sets. For our experiments we artificially create imbalance in the datasets by randomly subsampling the positive class. These imbalance subsets are fixed across all HMM approaches, and fixed separately across all deep-learning based approaches. 

\subsection{HMM-based Methods}

\subsubsection{HMMs and HMM-\textit{e}} For our experiments we adopt an ensemble size of 250 models per side, a sampling rate of 1\%, and 5 states in our HMMs. 

\subsubsection{\hmme + SVM} We train the downstream SVM using a radial basis function kernel, exploring an effective and simple model class in non-deep learning contexts.

\subsubsection{\hmme + NN} The architecture is similar to the baseline models in terms of number of layers and hidden dimensions; the classifier consists of four fully connected layers, starting with a hidden dimension of 512. Each layer incorporates batch normalization, ReLU activation, and dropout for regularization. The model is trained using binary cross-entropy loss, with the Adam optimizer and a learning rate scheduler. To address data imbalance, weighted sampling is applied during training.

\begin{table*}[tb]
\centering
\resizebox{\linewidth}{!}{
\begin{tabular}{ll|cc|cc|cc|cc|cc|cc}
\toprule
\multirow{2}{*}{\textbf{Dataset}} & \multirow{2}{*}{} & \multicolumn{2}{c|}{\textbf{CNN}} & \multicolumn{2}{c|}{\textbf{LSTM}} & \multicolumn{2}{c|}{\textbf{HMM}} & \multicolumn{2}{c|}{\textbf{\hmme}} & \multicolumn{2}{c|}{\textbf{\hmme + SVM}} & \multicolumn{2}{c}{\textbf{\hmme + NN}} \\
 & & \textit{AUC} & \textit{AP} & \textit{AUC} & \textit{AP} & \textit{AUC} & \textit{AP} & \textit{AUC} & \textit{AP} & \textit{AUC} & \textit{AP} & \textit{AUC} & \textit{AP} \\
\midrule
\multirow{2}{*}{\textbf{demo\_human\_or\_worm}} & 1:1 & 95.3  & 95.0  & 98.3 & 98.4  &  55.7&  55.6&  83.9&  84.6&  \textbf{99.2}&  \textbf{99.1} & 98.5  & 98.5  \\
 & 50:1 & 94.9  & 37.5 & 94.8  & 53.9 &  61.4&  14.0&  87.2&  27.5&  96.8&  73.2 & \textbf{98.4}  & \textbf{74.5} \\
\midrule
\multirow{2}{*}{\textbf{human\_nontata\_promoters}} & 1:1 & \textbf{89.5}  & \textbf{91.0} & 84.0  & 88.4 &  76.8&  66.1&  82.0&  73.2&  \textbf{89.5}&  84.0 & 88.0  & 86.2  \\
 & 50:1 & 81.0  & \textbf{13.4}  & 82.9  & 11.5  &  71.4&  4.1&  85.3&  9.2&  84.5&  11.4 & \textbf{90.5} & 11.8  \\
\midrule
\multirow{2}{*}{\textbf{human\_enhancers\_ensembl}} & 1:1 & 86.4  & 84.3  & 86.6  & 86.9 &  62.9&  57.8&  69.4&  69.9 & 85.0& 84.3 & \textbf{87.4} & \textbf{87.3} \\
 & 50:1 & \textbf{80.7} & 14.6  & 77.6 & 11.0 &  57.5&  2.4&  68.2&  6.4&  76.0&  \textbf{15.5} & 80.5  & 15.2 \\
\midrule
\multirow{2}{*}{\textbf{demo\_coding\_vs\_intergenomics}} & 1:1 & 94.4  & 93.7 & 90.2 & 89.8  &  76.3&  68.8&  85.1&  84.3&  95.5&  95.0 & \textbf{96.7}  & \textbf{96.6} \\
 & 50:1 & 88.2 & 32.6 & 88.4 & 23.1  &  74.6&  4.9&  87.9&  24.3&  85.0&  46.0 & \textbf{95.1} & \textbf{52.1} \\
\bottomrule
\end{tabular}}
\caption{Performance comparison of classification approaches across four datasets under balanced and imbalanced conditions. Metrics: AUC-ROC (\textit{AUC}; 0-100) and Average Precision (\textit{AP}; 0-100), $\times100$ for readability, with higher values indicating better performance. The HMM-\textit{e}+ SVM and HMM-\textit{e} + NN approaches consistently achieve superior performance, particularly in imbalanced datasets (50:1), demonstrating their robustness towards class imbalance.}
\label{tab:results}
\end{table*}

\subsection{Baselines} \label{sec:baselines}

Neural network approaches often face significant challenges when dealing with imbalanced classes in classification tasks. These models, driven by gradient-based optimization, tend to favor the majority class, leading to biased predictions and reduced performance on minority classes \cite{Wang2016TrainingDN}. The imbalance skews the learning process, causing the model to underfit the minority class, which is typically underrepresented in the loss function. Consequently, the network exhibits poor generalization, particularly in real-world applications where the minority class is critical. To mitigate the impact of class imbalances, several techniques have been proposed, including weighted loss functions, data augmentation, advanced sampling strategies, and hyperparameter tuning \cite{Buda_2018_advancedsampling, Cui_2019_CVPR_weightedloss, shwartz-hyperparametertuning, song2024_dataaugmentation}. However, despite the application of these ad-hoc methods, in practice, leveraging extremely imbalanced datasets using neural networks remains a significant challenge \cite{li2022autobalanceoptimizedlossfunctions}.

We compare our approaches against baseline Convolutional and LSTM based neural networks.

\subsubsection{CNN} We evaluate our approach against the baseline CNN in \cite{Gresova2022}. The model comprises of an embedding layer that maps the input tokens into dense vectors. These embeddings are processed through a sequence of 1-D convolutional layers with ReLU activations, batch normalization, and max-pooling. The model is optimized with the Adam optimizer and trained using the binary cross-entropy loss function for binary classification. The results of the CNN baseline are shown in Table \ref{tab:results}.

\subsubsection{LSTM} We implement a LSTM model for binary sequence classification. The architecture includes an embedding layer that maps input tokens to dense vectors, followed by a single LSTM layer with a hidden state dimension of 64. The LSTM output is fed into a dense layer with ReLU activation, and a sigmoid activation is applied to the final output. The model is trained using binary cross-entropy loss. The results of the CNN baseline are shown in Table \ref{tab:results}.

To address data imbalance during training, we employ weighted sampling during training for both the CNN and LSTM. Each sample in the training set is assigned a weight inversely proportional to its class frequency, giving higher weights to underrepresented classes. For balanced datasets, uniform weights are applied across all samples.

\subsection{Evaluation}
The choice of classification threshold (decision boundary) is critical as it impacts the trade-off between precision and recall, and directly influences metrics like classifier accuracy. In risk-sensitive domains, or in scenarios with ethical and regulatory considerations, this choice is further influenced by downstream factors that weigh the cost of false positives versus false negatives. For e.g., in fraud detection, a lower threshold will flag more fraud at the expense of increased false positives (user friction), but in healthcare, a lower threshold ensures patient safety, even at the expense of false alarms. At a fixed boundary, Accuracy can be a misleading metric, particularly in scenarios with data imbalance, for e.g., if anomalies constitute only 0.5\% of a dataset, a classifier can attain 99.5\% accuracy by labeling every sample identically (as normal). 

We use two popular techniques of evaluating classifiers across different decision boundaries: ROC and PR curves.

\subsubsection{AUC-ROC}
The Receiver Operating Characteristic (ROC) curve is used to understand the trade-off between the True Positive Rate (also known as Recall) and False Positive Rate at different classification boundaries/thresholds \cite{BRADLEY19971145}. The area under the curve (AUC) enables us to compare multiple ROC curves, with higher values suggesting better classifiers.  

\subsubsection{Average Precision (AP)} 
Precision-Recall (PR) curves are often used in settings with data imbalance, particularly when the positive class is rare~\cite{saito2015precision, davis2006relationship}. AP provides a way to condense the Precision-Recall (PR) curve into a singular metric; although it approximates the area under the PR curve (AUC-PR), it is not as optimistic in its computation as it does not interpolate, and instead computes the weighted average of precisions at the each threshold (weighted by change in recall) \cite{Zhang2009}.

\section{Results \& Discussion} \label{sec:results}

Table \ref{tab:results} presents a detailed performance evaluation of the various sequence classification approaches across 4 datasets in balanced (1:1) and imbalanced (50:1) class settings, reporting two metrics: AUC-ROC and AP.

\subsubsection{HMMs: Singleton vs Ensemble} We find that our \hmme ensemble approach consistently outperforms the singleton HMM approach, yielding a significant increase in both AUC and AP.  For example, on the \textit{demo\_human\_or\_worm} dataset, moving from a singleton HMM to an ensemble improves AUC from 55.7 to 83.9 and AP from 55.6 to 84.6. This trend continues under class imbalance: in the 50:1 class ratio setting,  the singleton HMM achieves 61.4 AUC and 14.0 AP, while \hmme achieves 87.2 and 27.5 AUC and AP respectively.

\subsubsection{Our Approach vs Baselines} While the baseline CNN and LSTM classifiers perform adequately on the balanced datasets, their efficacy diminishes with high class imbalance despite implementing weighted sampling approached to remedy this. The \hmme + SVM and \hmme + NN models consistently outperform other methods, particularly in the imbalanced data scenario (50:1). The combination of HMM ensembles with downstream classifiers offers significant advantages in capturing complex sequential relationships while mitigating the effects of class imbalance. 

In the few instances where the baseline CNN and LSTM methods outperform \hmme + SVM or \hmme + NN, the difference in performance is small. For example, for the human\_nontata\_promoters dataset, in the balanced data setting the CNN model has an AUC-ROC score of 89.5 and AP of 91.0, versus \hmme+SVM's 89.5 AUC and 84.0 AP.

\begin{figure}[bht]
    \centering
    \includegraphics[width=0.471\textwidth]{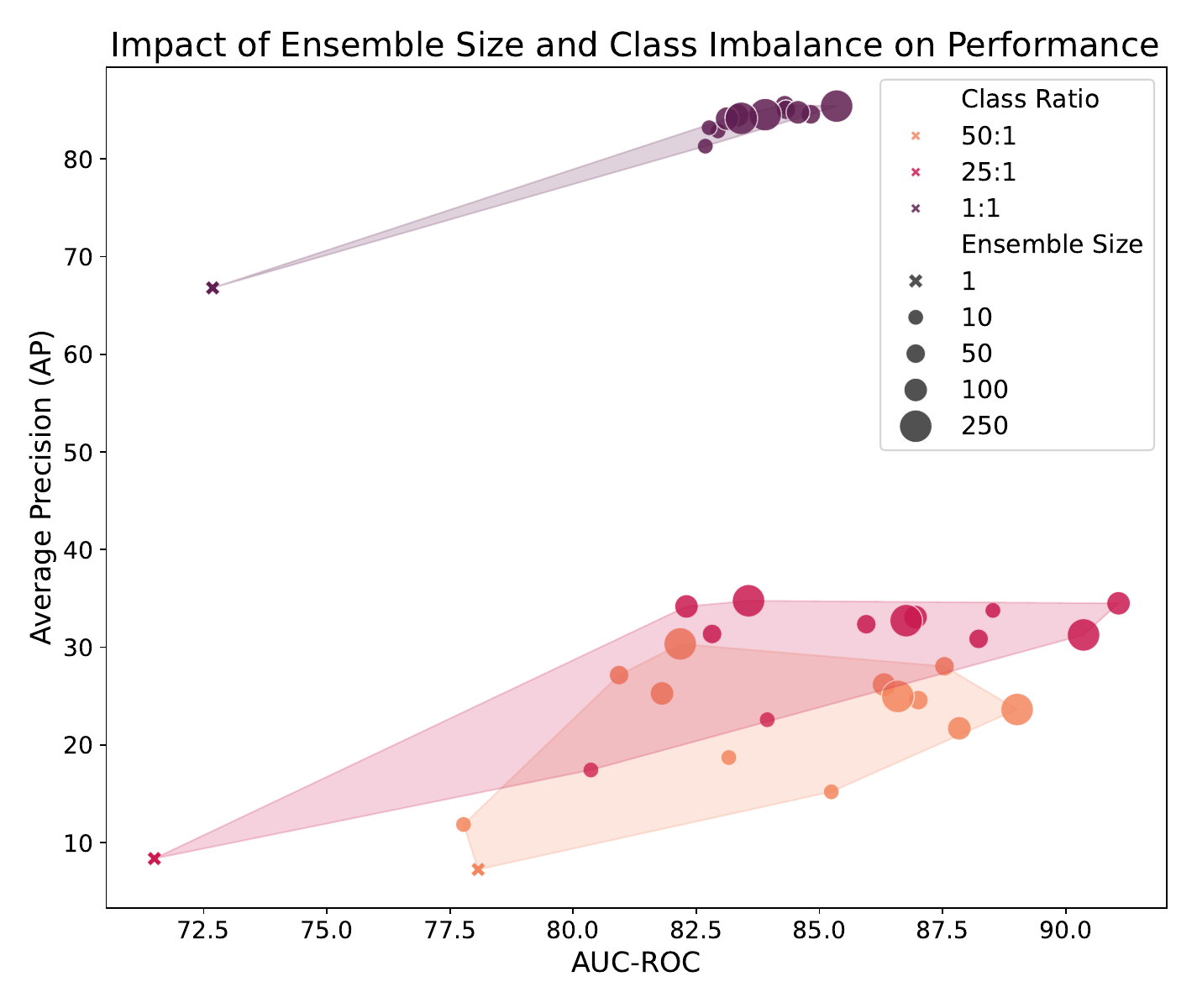}
    \caption{Impact of ensemble size and class imbalance on classifier performance. An increase in class imbalance causes significant decay in AP, as expected, along with a misleading increase in AUC-ROC. Larger ensemble sizes tend to perform better than smaller ones. }
    \label{fig:ensemble_performance}
\end{figure}

\subsubsection{Understanding Ensemble Parameters} \label{sec:ensemble_parameters}
While we present results with a fixed ensemble size in Table~\ref{tab:results}, the ensemble size parameters ($M$, $N$) and data subsample size ($s\%$) can be varied as discussed in Section~\ref{sec:formalization}. We examine how ensemble size and class imbalance impact HMM performance and summarize our findings in Figure~\ref{fig:ensemble_performance}. We conduct a detailed experiment on the demo\_human\_or\_worm dataset, training and evaluating models spanning three class imbalance settings (1:1, 25:1:, 50:1), five ensemble sizes (1, 10, 50, 100, 250), as well as varying data sample sizes (0.1\%, 1\%, 10\%).

Figure~\ref{fig:ensemble_performance} illustrates the impact of ensemble size and class imbalance on classifier performance. The ideal classifier would achieve AUC-ROC and AP of 100\%, and be positioned on the top-right of the plot. As expected, classifier performance decays with increased class imbalance. In general, larger ensembles outperform smaller ones, albeit only to a certain extent. Multiple values for the same (ensemble size, class ratio) configuration represent ensembles trained with different fractions of the dataset ($s$). We find that lower values of $s$ are essential at smaller ensemble sizes for creating diverse ensembles, whereas at larger ensemble sizes allow for larger values of $s$.

\section{Conclusion}
We introduce the \hmme framework, an ensemble approach for sequence classification that leverages HMMs. Our method demonstrates robust performance in scenarios with severe data imbalance, outperforming learned methods on Genomics Benchmark datasets. The \hmme framework is generalizable across problem definitions and has applications in domains spanning healthcare, finance, operations, and technology. Its compatibility with downstream modeling approaches (including both deep learning and traditional methods) makes it particularly useful in real-world settings, as it can be deployed in pipelines which may leverage additional data and domain knowledge. 

Key contributions include:
\begin{itemize}
    \item HMM-\textit{e}'s robustness to data imbalance
    \item Adaptability to multi-class settings
    \item Compatibility with downstream modeling approaches
    \item Inherent generative capabilities, lending it useful to synthetic data applications
\end{itemize}

Future work includes exploring approaches for identifying underperforming models in the ensemble and pruning or fine-tuning them. It also includes using unsupervised clustering on ensemble score features to identify classes of behavior, and leveraging the generative properties of HMMs for synthetic data generation.

\bibliography{aaai25}

\appendix

\begin{center}
\textbf{\large Supplementary Materials: Ensemble Methods for Sequence Classification with Hidden Markov Models}
\end{center}

\begin{table*}[t]
\centering
\begin{tabular}{ll|ccc|ccc}
\toprule
\multirow{2}{*}{\textbf{Method}} & \multirow{2}{*}{} & \multicolumn{3}{c|}{\textbf{AUC-ROC}} & \multicolumn{3}{c}{\textbf{Average Precision}} \\
 & & \textit{Mean} & \textit{Standard Deviation} & \textit{Standard Error} & \textit{Mean} & \textit{Standard Deviation} & \textit{Standard Error} \\
\midrule
\multirow{2}{*}{\textbf{CNN}} & 1:1 & 93.92 & 0.22 & 0.12 & 92.48 & 0.48  & 0.27 \\
& 50:1  & 88.16  & 0.46 & 0.26 & 30.35 & 2.69  & 1.55 \\
\midrule
\multirow{2}{*}{\textbf{LSTM}} & 1:1 & 91.99 & 1.28 & 0.73 & 91.54 & 1.48 & 0.85 \\
 & 50:1 & 88.30 & 2.67 & 1.54 & 23.36 & 7.76 & 4.48 \\
\midrule
\multirow{2}{*}{\textbf{HMM}} & 1:1 & 76.62 & 0.22 & 0.13 & 69.75 & 0.43 & 0.25 \\
 & 50:1 & 77.21 & 1.43 & 0.82 & 5.03 & 0.50 & 0.29 \\
\midrule
\multirow{2}{*}{\textbf{HMM-e}} & 1:1 & 85.38 & 0.05 & 0.03 & 84.40 & 0.23 & 0.13 \\
 & 50:1 & 87.91 & 0.28 & 0.16 & 25.21 & 1.33 & 0.77\\
\midrule
\multirow{2}{*}{\textbf{HMM-e + SVM}} & 1:1 & 95.24& 0.18& 0.11& 94.70& 0.22& 0.13\\
 & 50:1 & 86.08& 1.67 & 0.96 & 44.61 & 2.10 & 1.21 \\
\midrule
\multirow{2}{*}{\textbf{HMM-e + NN}} & 1:1 & 96.72 & 0.01 & 0.00 & 96.66 & 0.02 & 0.01 \\
 & 50:1 & 94.71 & 0.17 & 0.98 & 52.03 & 0.5 & 0.28 \\
\bottomrule
\end{tabular}
\caption{Mean, standard deviation, and standard error of mean for different methods on the \textit{demo\_coding\_vs\_intergenomic\_seqs} dataset under balanced (1:1) and imbalanced (50:1) conditions. Results are presented across 3 training runs of all methods. Metrics: AUC-ROC (0-100) and Average Precision (0-100).}
\label{tab:uncertainty}

\label{tab:std_se}
\end{table*}

\section{Quantifying Uncertainty} \label{sec:uncertainty_quantification}

In Table \ref{tab:uncertainty}, we present results from all methods explored in our study, with uncertainty measures, on the dataset \textit{demo\_coding\_vs\_intergenomic\_seqs}. These findings are computed both under balanced (1:1) and imbalanced (50:1) conditions, across 3 training runs. 

We find, as expected, that the standard deviation of both AUC-ROC and average precision rises across all methods in the imbalanced setting compared to the balanced setting. \hmme, \hmme+SVM, and \hmme+NN have less variance in their performance than both the CNN and LSTM learned baselines, across data imbalance settings.

\section{Datasets}

To demonstrate our approach's effectiveness, we leverage an open-source genomics benchmark (Gresova et al. 2022) that consists of eight datasets for different classification problems. We provide results on four datasets included in the benchmark. 

All datasets are split by the benchmark authors into train and test sets. For our experiments we artificially create imbalance in the datasets by randomly subsampling the positive class.

\subsubsection{human\_nontata\_promoters} Originally drawn from (Umarov and Solovyev 2017), this dataset consists of 36,131 sequences with a median length of 251. These sequences are drawn from the human genome and fall into one of two well-known classes of promoter genes -- TATA or non-TATA. 

\subsubsection{human\_enhancers\_ensembl} This dataset is documented in (Andersson et al. 2014) and consists of 154,842 sequences with median length 269. Positive sequences are human enhancers and negative sequences are randomly selected, non-overlapping segments from human genome GRCh38. 

\subsubsection{demo\_human\_or\_worm, \\demo\_coding\_vs\_intergenomic\_seqs} These two datasets are computationally generated by the authors of (Gresova et al. 2022) by sampling randomly from human and worm genomes. Both consist of 100,000 sequences with median length 200. In \textit{demo\_human\_or\_worm} the classes correspond to the organism from which the transcript is sampled, and in \textit{demo\_coding\_vs\_intergenomic\_seqs} the classes correspond to whether the transcript codes for proteins or does nothing.

\section{Hyperparameter Setting} \label{sec:hyperparam_supplementary}
\subsubsection{CNN} We adopt the same CNN hyperparameters as in the baseline included in (Gresova et al. 2022). The CNN model comprises of three convolutional layers with 16, 8, and 4 filters, respectively, each with a kernel size of 8 and a stride of 1. Batch normalization and ReLU activation follow each convolution, with 2x max-pooling applied after each block. The model includes two dense layers, with the first layer containing 512 units. The default learning rate of 0.001 set by the Adam optimizer is used. The output layer utilizes a sigmoid activation. We use a binary cross-entropy loss.
\subsubsection{LSTM} The LSTM baseline model consists of a single LSTM layer with a hidden dimension of 64, followed by two dense layers (256 units and an output layer). The network includes an embedding layer with a configurable embedding dimension (set to 100). The Adam optimizer is used with a learning rate of 0.001. For binary classification, a sigmoid activation and binary cross-entropy loss are used. 
\subsubsection{HMM} We perform a search for the number of states underlying our HMMs, ranging over [2, 7]. From an AUC and AP perspective we find that the number of states makes little difference to performance, so we choose 5 as it sits in the middle of the range.
\subsubsection{\hmme} We perform a search over three hyperparameters: ensemble size, subset factor, and number of states per HMM. For ensemble size we try values in [10, 50, 100, 250, 500]. For subset factor we try values in [0.1\%, 1\%, 2\%, 3\%, 5\%, 10\%]. For number of states we try values in the range [2, 7]. We find ensemble size, then subset factor, then number of states per HMM to be the most important hyperparameters for performance. We select our final setting -- ensemble size of 250, subset factor of 1, and number of states of 5 -- to maximize AUC and AP.
\subsubsection{\hmme + SVM} We use almost all default hyperparameters for our SVM and perform no hyperparameter search. The only non-standard hyperparameter we use is to set \textit{probability=True} which allows us to compute log-likelihoods of sequences under the model after training.
\subsubsection{\hmme + NN} The MLP classifier architecture follows that of the baseline CNN but with fully connected layers in place of the Convolutional layers. It features four fully connected layers with hidden dimensions of 512, 256, and 128, employing batch normalization and dropout (0.25) after each ReLU activation. The model is optimized using the Adam optimizer with a learning rate of 0.001 and a batch size of 64. The output dimension is set to 1 for binary classification

\section{Hardware and Software Stack} \label{sec:hardware_software}

Our experiments are performed on an AWS \textit{r5.24xlarge} EC2 instance featuring 96 virtual CPUs and 768 GB of memory. Due to the lightweight nature of the models trained, we do not have to leverage GPU acceleration. The environment is configured with Ubuntu 20.04 LTS as the operating system, and we use Python version 3.8.10. Aside from standard machine-learning libraries like Pandas, NumPy, Pytorch, Scikit-Learn, and Tensorflow, we also use HMMLearn to train our HMMs, as well as the Github repository linked in (Gresova et al. 2022) to load our datasets.

\section{Pseudocode} \label{sec:pseudocode}

We include detailed pseudocode for our method which should provide sufficient guidance for anyone looking to replicate our results. This pseudocode covers all aspects of our approach, including: 
\begin{enumerate}
    \item Algorithm \ref{alg: hmm-e}: Details the basic framework for training our \hmme approach. A fuller description can be found in the paper in Algorithm 1.
    \item Algorithm \ref{alg: evaluate}: Evaluation code for computing metrics on trained models.
    \item Algorithm \ref{alg: get_feature_vectors}: Details the approach for constructing feature vectors for training downstream classifiers, given trained models.
    \item Algorithm \ref{alg: train_svm}: Code for training SVMs on \hmme feature vectors.
    \item Algorithm \ref{alg: train_mlp_classifier}: Code for training neural networks on \hmme feature vectors.
\end{enumerate}

\begin{algorithm}[h] 
\caption{HMM-e Algorithm: Main Pipeline}
\label{alg: hmm-e}
\begin{algorithmic}[1]
\REQUIRE Dataset name $dataset$, class ratio $class\_ratio$, $ensemble\_size$, $subset\_factor$
\STATE (positive\_train, negative\_train, positive\_val, negative\_val) = \textit{load\_dataset}(dataset, class\_ratio)
\STATE training\_jobs =\textit{ get\_training\_jobs}(positive\_train, negative\_train, ensemble\_size, subset\_factor)
\STATE trained\_models = \textit{train\_models}(training\_jobs, ensemble\_size)
\STATE \textit{save\_models}(trained\_models)
\STATE results = \textit{evaluate\_models}(trained\_models, negative\_val, positive\_val)
\STATE \textit{store\_results}(results)
\end{algorithmic}
\end{algorithm}

\begin{algorithm}[h]
\caption{evaluate\_models}
\label{alg: evaluate}
\begin{algorithmic}[1]
\REQUIRE Trained models $trained\_models$, normal validation sequences $negative\_val$, anomalous validation sequences $positive\_val$
\STATE true\_classes = []
\STATE scores = []
\FOR{sequence in negative\_val + positive\_val}
    \STATE scores.append(\textit{composite\_score}(sequence, trained\_models))   
    \hfill \textit{// See Eqn. 4 for composite score.}
    \IF{sequence in negative\_val}
        \STATE true\_classes.append(0)
    \ELSE
        \STATE true\_classes.append(1)
    \ENDIF
\ENDFOR
\STATE auc\_roc = \textit{sklearn.metrics.roc\_auc\_score}(true\_classes, scores)
\STATE ap = \textit{sklearn.metrics.average\_precision\_score}(true\_classes, scores)
\RETURN auc\_roc, ap
\end{algorithmic}
\end{algorithm}

\begin{algorithm}[h]
\caption{generate\_feature\_vectors}
\label{alg: get_feature_vectors}
\begin{algorithmic}[1]
\REQUIRE Trained models $trained\_models$, anomalous sequences $anomalous$, normal sequences $normal$
\STATE F = []
\FOR{sequence in anomalous + normal}
    \STATE f = []
    \FOR{model in trained\_models}
        \STATE f.append(model.\textit{log\_likelihood}(sequence))
    \ENDFOR
    \STATE F.append(f)
\ENDFOR
\RETURN F
\end{algorithmic}
\end{algorithm}

\begin{algorithm}[h]
\caption{train\_svm}
\label{alg: train_svm}
\begin{algorithmic}[1]
\REQUIRE Trained models $trained\_models$, anomalous sequences $positive\_train$, normal sequences $negative\_train$
\STATE F\_train = \textit{generate\_feature\_vectors}(trained\_models, positive\_train, negative\_train)
\STATE true\_classes = [0 if seq in negative\_train else 1 for seq in negative\_train + positive\_train]
\STATE svm\_model = \textit{sklearn.svm.SVC().fit}(F\_train, true\_classes)
\RETURN svm\_model
\end{algorithmic}
\end{algorithm}

\begin{algorithm}[h]
\caption{train\_mlp\_classifier}
\label{alg: train_mlp_classifier}
\begin{algorithmic}[1]
\REQUIRE Training dataset $train\_dataset$, Testing dataset $test\_dataset$, Sample weights $train\_weights$, Input dimension $input\_dim$, Hidden dimension $hidden\_dim$, Learning rate $learning\_rate$, Batch size $batch\_size$
\STATE data\_module = DataLoader(train\_dataset, test\_dataset, batch\_size, train\_weights)
\STATE model = MLPClassifier(input\_dim, hidden\_dim, output\_dim=1, learning\_rate)
\STATE trainer = Trainer(max\_epochs=16)
\STATE trainer.fit(model, datamodule=data\_module)
\RETURN model
\end{algorithmic}
\end{algorithm}

\end{document}